\documentclass[journal]{IEEEtran}
\IEEEoverridecommandlockouts
\usepackage{cite}
\usepackage{amsmath,amssymb,amsfonts}
\usepackage{algorithmic}
\usepackage{graphicx}
\usepackage{textcomp}
\usepackage{multirow}
\usepackage{xcolor}
\usepackage{tabularx}
\usepackage{booktabs}
\usepackage{breqn}  
\usepackage{lipsum}
\usepackage{stfloats}
\usepackage{hyperref}
\usepackage{subcaption}
\usepackage{amssymb}
\usepackage[ruled,vlined]{algorithm2e}
\usepackage{bm}
\usepackage{subcaption}
\usepackage{bbm}

\def\BibTeX{{\rm B\kern-.05em{\sc i\kern-.025em b}\kern-.08em
    T\kern-.1667em\lower.7ex\hbox{E}\kern-.125emX}}
    
\usepackage{float} 

\begin{document}

\markboth{This Article Has Been Published in \textbf{IEEE Transactions on Emerging Topics in Computational Intelligence}}%
	    {This Article Has Been Published in \textbf{IEEE Transactions on Emerging Topics in Computational Intelligence}}

\title{ADAST: Attentive Cross-domain EEG-based Sleep Staging Framework with Iterative Self-Training}

\author{Emadeldeen Eldele, Mohamed Ragab, Zhenghua Chen, Min Wu, Chee-Keong Kwoh, Xiaoli Li \\and Cuntai Guan~\IEEEmembership{Fellow,~IEEE}

\thanks{Emadeldeen Eldele, Chee-Keong Kwoh and Cuntai Guan are with the School of Computer Science and Engineering, Nanyang Technological University, Singapore (E-mail: \{emad0002, asckkwoh, ctguan\}@ntu.edu.sg).}
\thanks{Mohamed Ragab and Xiaoli Li are with Institute for Infocomm Research (I$^2$R), Centre for Frontier Research (CFAR), Agency of Science, Technology and Research (A$^*$STAR), Singapore, and also with the School of Computer Science and Engineering at Nanyang Technological University, Singapore (E-mail: mohamedr002@e.ntu.edu.sg, xlli@i2r.a-star.edu.sg).}
\thanks{Zhenghua Chen is with the Institute for Infocomm Research (I$^2$R) and the Centre for Frontier AI Research (CFAR), Agency for Science, Technology and Research (A$^*$STAR), Singapore (E-mail: chen0832@e.ntu.edu.sg).}
\thanks{Min Wu is with the Institute for Infocomm Research (I$^2$R), Agency for Science, Technology and Research (A$^*$STAR), Singapore (E-mail: wumin@i2r.a-star.edu.sg).}
\thanks{First and second authors are supported by A$^*$STAR SINGA Scholarship.}
\thanks{Min Wu is the corresponding author.}

\thanks{\copyright 2022 IEEE.  Personal use of this material is permitted.  Permission from IEEE must be obtained for all other uses, in any current or future media, including reprinting/republishing this material for advertising or promotional purposes, creating new collective works, for resale or redistribution to servers or lists, or reuse of any copyrighted component of this work in other works.}

}

\maketitle

\begin{abstract}
Sleep staging is of great importance in the diagnosis and treatment of sleep disorders. Recently, numerous data-driven deep learning models have been proposed for automatic sleep staging. 
They mainly train the model on a large public labeled sleep dataset and test it on a smaller one with subjects of interest. However, they usually assume that the train and test data are drawn from the same distribution, which may not hold in real-world scenarios. Unsupervised domain adaption (UDA) has been recently developed to handle this domain shift problem. However, previous UDA methods applied for sleep staging have two main limitations. First, they rely on a totally shared model for the domain alignment, which may lose the domain-specific information during feature extraction. Second, they only align the source and target distributions globally without considering the class information in the target domain, which hinders the classification performance of the model while testing. In this work, we propose a novel adversarial learning framework called ADAST to tackle the domain shift problem in the unlabeled target domain. First, we develop an unshared attention mechanism to preserve the domain-specific features in both domains. Second, we design an iterative self-training strategy to improve the classification performance on the target domain via target domain pseudo labels. We also propose dual distinct classifiers to increase the robustness and quality of the pseudo labels. The experimental results on six cross-domain scenarios validate the efficacy of our proposed framework and its advantage over state-of-the-art UDA methods. 
The source code is available at \href{https://github.com/emadeldeen24/ADAST}{https://github.com/emadeldeen24/ADAST}.

\end{abstract}


\section{Introduction}
Sleep stage classification is crucial to identifying sleep problems and disorders in humans. This task refers to the classification of one or many different signals including electroencephalography (EEG), electrocardiogram (ECG), electrooculogram (EOG), and electromyogram (EMG) into one of five sleep stages, namely, wake (W), rapid eye movement (REM), non-REM stage 1 (N1),  non-REM stage 2 (N2), and non-REM stage 3 (N3). 
For EEG recordings, they are usually split into 30-second segments, where each segment is classified manually into one of the above stages by specialists \cite{bands}. Despite being mastered by many specialists, the manual annotation process is tedious and time-consuming, especially with the large amount of collected EEG data. 

\begin{figure}[t]
    \centering
    \includegraphics[width=\columnwidth]{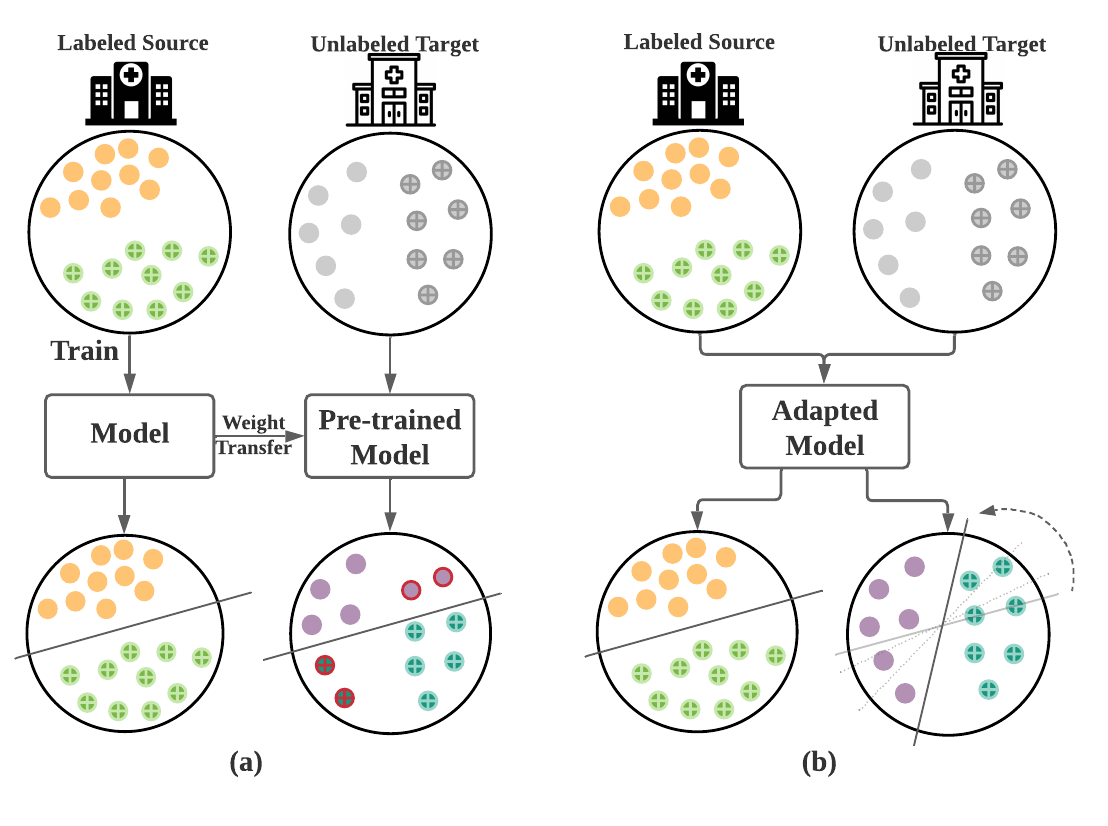}
    \caption{(a) Direct Transfer (DT) fails due to the domain shift, and (b) Domain Adaptation (DA) solves the domain shift problem.}
    \label{fig:dt_da_comparison}
\end{figure}

In recent years, numerous data-driven deep learning approaches have been developed, relying on the availability of a massive amount of labeled data for training. Therefore, many deep learning methods have been proposed to perform sleep staging automatically \cite{deepsleepnet,tnnls_cnn_paper,seqsleepnet,attnSleep_paper}.
These methods implemented different network structures to process EEG data and trained proper classification models to achieve good performance while testing. Since these methods were able to achieve decent performance, it was expected to be a step forward to reduce the reliance on the manual scoring process. However, many sleep labs were found to keep relying on manually scoring EEG data \cite{attentive_sleep_Staging,phan2020towards}.
The main reason is the high variation between the public training data and the data generated in the sleep labs. These variations can occur due to several factors, \textit{e.g.}, different measuring locations on the skull and different sampling rates of measuring devices. This is well-known as the \textit{domain shift} problem, i.e., the training (\emph{source}) and testing (\emph{target}) data have different distributions. 
As shown in Fig.~\ref{fig:dt_da_comparison}(a), directly applying the source-pretrained model (i.e., Direct Transfer) on the target data may not well-classify the target domain data due to the domain shift. Consequently, these models suffer significant performance degradation when trained on public datasets and tested on the sleep labs data. In addition, it is difficult for these labs to annotate large enough EEG datasets to re-train the models.

A typical solution for the above issues is to employ transfer learning approaches \cite{phan2020towards,channel_mismatch}. For instance, Phan \textit{et al.} \cite{phan2020towards} applied transfer learning from a large dataset to a different and relatively smaller one. They first pre-trained the model on a large dataset and then fine-tuned it on a smaller dataset. Similarly, the authors in \cite{channel_mismatch} studied the channel mismatch problem while transferring the knowledge from one dataset to another. However, these transfer learning methods require the availability of labeled data from the target domain to fine-tune the model. In reality, the target domain may be completely unlabeled, and it is thus impractical to fine-tune the models.

Unsupervised Domain Adaptation (UDA) is a special scenario of transfer learning that aims to minimize the mismatch between the source and target distributions without using any target domain labels. As shown in Fig.~\ref{fig:dt_da_comparison}(b), UDA aims to use both the labeled source domain along with the unlabeled target domain to train the model in a way that allows it to perform well on both source and target domains.
So far, limited studies have investigated UDA in the context of sleep stage classification. For instance, Chambon \textit{et al.} \cite{da_sleep} improved the feature transferability between source and target domains using optimal transport domain adaptation. In addition, Nasiri \textit{et al.} \cite{attentive_sleep_Staging} used adversarial training-based domain adaptation to improve the transferability of features.
However, these methods still suffer from the following limitations. First, they rely on totally shared models (i.e., same architectures with same weights) to extract features from both source and target domains. This may lose the domain-specific features for both source and target domains, which can be harmful to the classification task on the target domain. Second, these approaches only align the global distribution between source and target domains without considering the mismatch of the fine-grained class distribution between the domains. As such, target samples belonging to one class can be misaligned with an incorrect class in the source domain.

To tackle the aforementioned challenges, we propose an \textbf{A}dversarial \textbf{D}omain \textbf{A}daptation framework based on preserving attention mechanism and iterative \textbf{S}elf-\textbf{T}raining strategy (\textbf{ADAST}) for a single channel EEG-based sleep stage classification. Specifically, we first propose a domain-specific attention module to preserve both the source-specific and the target-specific features. This helps to keep the main characteristics of both domains to improve adversarial learning. Second, we propose an iterative self-training strategy to well-classify the fine-grained distribution of the unlabeled target domain using a target domain pseudo labels supervision. Hence, we can adapt the classification decision boundaries according to the target domain classes. Moreover, we design distinct dual classifiers to improve the robustness of target domain pseudo labels. 

The main contributions of this work are summarized as follows:
\begin{itemize}
    \item We propose a novel cross-dataset sleep staging framework that integrates iterative self-training with adversarial learning. Therefore, our framework can effectively classify the fine-grained distribution of the unlabeled target sleep data.
    
    \item ADAST utilizes an unshared domain-specific attention module to preserve the key features in both source and target domains during adaptation, which improves the adversarial training and boosts the classification performance on the target domain.

    \item We design distinct dual classifiers to improve the robustness of the generated pseudo labels in self-training. We also add a similarity constraint on their weights to push them from being identical.
    
    \item Extensive experiments demonstrate that our ADAST achieves superior performance for cross-domain sleep stage classification against state-of-the-art UDA methods.
\end{itemize}

\section{Related Works}

\subsection{Sleep Stage Classification}
\label{subsec:ssc}

Automatic sleep staging with single-channel EEG has been widely studied in the literature. In particular, deep learning-based methods \cite{deepsleepnet,attnSleep_paper,seqsleepnet} have shown great advances through end-to-end feature learning. These methods design different network structures to extract the features from EEG data and capture the temporal dependencies.

Several studies explored convolutional neural networks (CNN) for feature extraction from EEG data. For example, Supratak \textit{et al.} \cite{deepsleepnet} proposed two CNN branches to extract different frequency features in EEG signals. The same CNN architecture was also adopted by Mousavi \textit{et al.} \cite{mousavi2019sleepeegnet}. Li \textit{et al.} \cite{cnn_se_Sleep} proposed to adopt CNN in addition to a squeeze and excitation block to extract the features from multi-epoch EEG data. Eldele \textit{et al.} \cite{attnSleep_paper} developed a multi-resolution CNN with an adaptive features recalibration to extract representative features. Additionally, Qu \textit{et al.} \cite{residual_attn} proposed multiple residual CNN blocks to learn features mappings. The above methods further handled the temporal dependencies in EEG epochs. They either used recurrent neural networks (RNNs), such as Long Short-Term Memory (LSTM) as in \cite{deepsleepnet,mousavi2019sleepeegnet}, or adopted the multi-head self-attention approach as in \cite{attnSleep_paper,residual_attn}.

Different from relying on CNNs, researchers proposed different ways to handle EEG data.
For example, Phan \textit{et al.} \cite{seqsleepnet} designed an end-to-end hierarchical RNN architecture. It consists of an attention-based recurrent layer to handle the short-term features within EEG epochs, besides a recurrent layer to capture the epoch-wise features.
In addition, Phan \textit{et al.} \cite{xsleepnet} used both raw EEG signal and its time-frequency image to design a joint multi-view learning from both representations. 
Also, Jia \textit{et al.} \cite{graphsleepnet} proposed a graph-based approach for sleep stage classification, where graph and temporal convolutions were utilized to extract spatial features and capture the transition rules, respectively. 
Finally, Neng \textit{et al.} \cite{ccrrsleepnet} handled the EEG data in three levels: frame, epoch, and sequence levels to extract a mixture of features that would improve the classification performance.
Despite the success of these methods in handling complex EEG data, their performance for cross-domain (e.g., cross-dataset) sleep stage classification is limited due to the domain shift issue. Therefore, many researches were directed to adopt transfer learning approaches to handle this issue.

\begin{figure*}
\centering
\includegraphics[width=\textwidth]{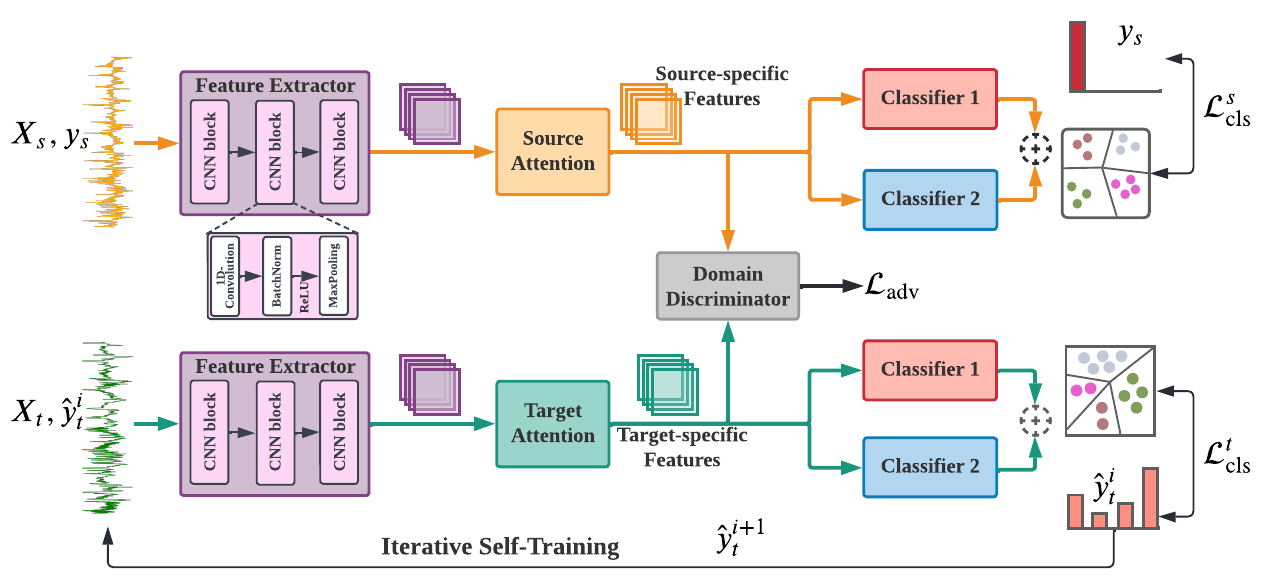}

\caption{The overall architecture of the proposed ADAST framework. The shared feature extractor consists of three convolutional blocks, where each block contains 1D-convolution, batch normalization, non-linear ReLU activation, and MaxPooling. The two classifiers share the same architecture, but we apply a similarity constraint on their weights to push them from being identical to each other (best viewed in colors, as blocks with similar colors represent shared components).}
\label{Fig:end-to-end}
\end{figure*}

\subsection{Transfer Learning for Sleep Staging}

Some works studied the problem of personalized sleep staging to improve the classification accuracy for individual subjects within the same dataset using transfer learning~\cite{personalized_1,personalized_2}. 
For a dataset with two-night recordings for each subject, these works pretrained the model by excluding the two nights of the test subject. Next, the first night is applied for fine-tuning the model and the second night is used for evaluation.
However, few works have been proposed for the cross-dataset scenario, i.e., training a model on subjects from one dataset and testing on different subjects from another dataset. Phan \textit{et al.} \cite{phan2020towards} studied the data-variability issue with the availability of large source dataset, and different labeled but insufficient target dataset. They trained their model on the source dataset and fine-tuned it on the smaller target dataset. With a similar problem setting, Phan \textit{et al.} \cite{channel_mismatch} proposed to use deep transfer learning to overcome the problem of channel mismatch between the two domains.

These methods require large corpus source datasets to increase their generalization ability and a labeled target dataset to fine-tune their models. Unsupervised domain adaptation (UDA) approaches were proposed to address these issues by aligning the features from different domains. These approaches can be categorized as discrepancy-based approaches and adversarial-based approaches. The discrepancy-based approaches such as Maximum Mean Discrepancy (MMD) \cite{mmd} and CORrelation ALignment (CORAL) \cite{coral}, attempt to minimize the distance metric between the source and target distributions. 
One the other hand, adversarial-based approaches mimic the adversarial training proposed in the generative adversarial network (GAN) \cite{goodfellow_gan}. This approach is more popular in previous sleep staging works. For example, Zhao \textit{et al.} \cite{zhao2021unsupervised} proposed using adversarial UDA with a domain discriminator and multiple classifiers fed from the different feature extractor layers. Nasiri \textit{et al.} \cite{attentive_sleep_Staging} used adversarial training along with local and global attention mechanisms to extract the transferable individual information. Yoo \textit{et al.} \cite{yoo2021transferring} proposed using adversarial domain adaptation with three discriminators; one for global alignment and two for stage and subject discrimination.

Differently, we enhance the adversarial training process by preserving the domain-specific features through domain-specific attention. Besides, we consider the fine-grained domain classes with the iterative self-training strategy, which deploys the target pseudo labels to improve the classification performance in the unseen target domain.

\section{Method}

\subsection{Preliminaries}
In this work, we focus on the problem of unsupervised cross-domain adaptation for EEG-based sleep staging. 
In this setting, we have access to a labeled source dataset ${X}_s= \{(\mathbf{x}_s^i,y_s^i)\}_{i=1}^{n_s}$  of $n_s$ labeled samples, and an unlabeled target dataset ${X}_t= \{(\mathbf{x}_t^j)\}_{j=1}^{n_t}$ of $n_t$ samples. The source and target domains are sampled from source distribution $P_s(X_s)$ and target distribution $P_t(X_t)$ respectively, such that these distributions are different (i.e., $P_s \neq P_t$). Both domains share the same label space $Y=\{1,2, \dots K\}$, where $K$ is the number of classes (i.e., sleep stages). The domain adaptation scenario aims to transfer the knowledge from a labeled source domain to a domain-shifted unlabeled target domain. In the context of EEG data, both $\mathbf{x}_s^i$ and $\mathbf{x}_t^i$ $\in  \mathbb{R}^{1 \times T}$, where the number of electrodes/channels is $1$ since we use single-channel EEG data, and $T$ represents the number of timesteps in the 30-second EEG epochs.

\subsection{Overview}
As shown in Fig.~\ref{Fig:end-to-end}, our proposed framework consists of three main components, namely domain-specific attention, adversarial training, and dual classifier-based iterative self-training.
First, domain-specific attention plays an important role in refining the extracted features so that each domain preserves its key features. 
Second, the adversarial training step leverages a domain discriminator to align the source and target features. Particularly, the domain discriminator network is trained to distinguish between the source and target features while the feature extractor is trained to confuse the domain discriminator by generating domain invariant features.
Finally, the iterative self-training strategy utilizes the target domain pseudo labels to adapt the classification decision boundaries according to the target domain classes. The dual classifiers are incorporated to improve the quality and robustness of the pseudo labels. Further details about each component will be provided in the following subsections.

\subsection{Domain-specific Attention}
Our proposed framework extracts domain invariant features by using a shared CNN-based feature extractor, i.e., $F_s(\cdot) = F_t(\cdot) = F(\cdot)$. 
Unlike the totally unshared architectures that require an extra pretraining step and are usually harder to converge, the shared feature extractor allows end-to-end training, besides being easier to converge. Therefore, most UDA algorithms adopted this shared design \cite{uda_survey}.
However, relying solely on this shared architecture may not be able to preserve the key features of each domain \cite{DS_paper1,DS_paper2}. 
The reason is that the high-dimensional features may contain information that distinguishes source and target domains, and are relevant for predicting the label \cite{ds_reason1}. Throughout the training, the model tries to remove these features to reduce the difference between source and target distributions and make the features domain-invariant, but this also affects the classification performance on each separate domain.
As the classification mainly relies on the available source domain labels, the learned classifier will be more biased toward the source domain \cite{ds_reason2}.
Hence, we propose an unshared attention module to learn both domain-invariant and domain-specific features jointly in our proposed framework.

\begin{figure} 
    \begin{center}
    \includegraphics[width=\columnwidth]{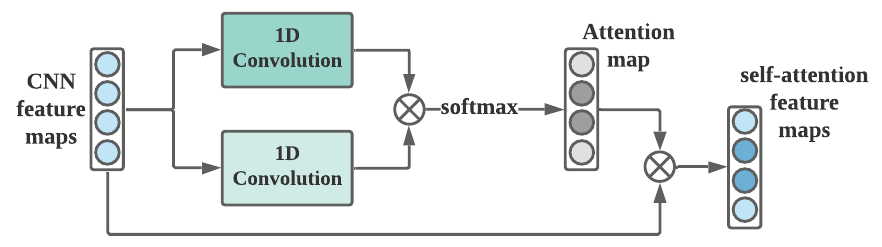}
    \end{center}
    \caption{Design of domain-specific attention module.}
    \label{fig:self-attn}
\end{figure}

For each position in the feature space, the attention module calculates the weighted sum of the features at all positions with a little computational cost.
Thus, the features at each location have fine details that are coordinated with fine details in distant portions of the features.
Formally, given an input source sample $\mathbf{x}_s \in \mathbb{R}^{1 \times T}$ that is passed through the feature extractor to generate the source features, i.e., $F(\mathbf{x}_s) = (\mathbf{f}_{s1}, \dots, \mathbf{f}_{sl})\in \mathbb{R}^{d \times l}$, where $d$ is the number of CNN channels, and $l$ is the length of the features. 
Inspired by \cite{sagan}, we deploy a convolutional attention mechanism as shown in Fig.~\ref{fig:self-attn}. 
The attention operation starts by obtaining new representation for the features at each position by using two 1D-convolutions, i.e., $H_1$ and $H_2$. Specifically, given $\mathbf{f}_{si}, \mathbf{f}_{sj} \in \mathbb{R}^{d}$, which are the feature values at the positions $i$ and $j$, they are transformed into $\mathcal{Z}_{si} = H_1(\mathbf{f}_{si})$ and $\mathcal{Z}_{sj} = H_2(\mathbf{f}_{sj})$.
The attention scores are calculated as follows.

\begin{equation}
    \mathcal{V}_{ji} = \frac{\exp (\mathcal{Z}_{si}^\top \mathcal{Z}_{sj})}{\sum_{k=1}^{l} \exp(\mathcal{Z}_{sk}^\top \mathcal{Z}_{sj})}. \label{eqn:attn_map} 
\end{equation}

Here, the attention score $\mathcal{V}_{ji}$ indicates the extent to which $j^{th}$ position attends to the $i^{th}$ position in the feature map.
The output of the attention layer is $\mathcal{O}_{s} = (\mathbf{o}_{s1}, \dots, \mathbf{o}_{sj}, \dots \mathbf{o}_{sl}) \in \mathbb{R}^{d \times l}$, where
\begin{equation}
\mathbf{o}_{sj} = \sum_{i=1}^{l} \mathcal{V}_{ji} {\mathbf{f}_s}_i.
\label{eqn:attn_out}
\end{equation}
We denote the attention process in Equations~\ref{eqn:attn_map} and \ref{eqn:attn_out} as $A(\cdot)$, such that $\mathcal{O}_s = A_s(F(\mathbf{x}_s))$. 
The same process applies to the target domain data flow to train $A_t$.

\subsection{Adversarial Training}
Given the learned source and target representations that preserve the domain-specific features, adversarial training is employed to align the source and target domains. Inspired by the generative adversarial network (GAN) \cite{goodfellow_gan}, we aim to solve a minimax objective between the feature extractor and domain discriminator. Specifically, the domain discriminator is trained to classify between the source and target features, while the feature extractor tries to generate indistinguishable representations for both source and target domains. By doing so, the classifier trained on the source domain can generalize well on the target domain. However, with the minimax objective, the discriminator can saturate quickly, resulting in a gradient vanishing problem \cite{adda}. To address this issue, we train our model using a standard GAN loss with inverted labels \cite{goodfellow_gan}. Formally, the domain discriminator, $D$, classifies the input features to be either from the source or target domain.
Thus, $D$ can be optimized using a standard cross-entropy loss with the labels indicating the domain of the data point.
The objective of this operation $\mathcal{L}_{D}$ can be defined as:
\begin{align}
\min_D \mathcal{L}_{\mathrm{D}}= 
&-\mathbb{E}_{\mathbf{x}_{s} \sim P_{s}}[\log D(A_s(F(\mathbf{x}_{s})))] \nonumber \\
&-\mathbb{E}_{\mathbf{x}_{t} \sim P_{t}}[\log (1-D(A_t(F(\mathbf{x}_{t}))))],
\label{eqn:train_disc}
\end{align}
where $\mathcal{L}_{\mathrm{D}}$ is used to optimize the domain discriminator separately so that it discriminates the source and target features.
On the other hand, the feature extractor and the domain-specific attention are trained to confuse the discriminator by mapping the target features to be similar to the source ones. The objective function can be described as:
\begin{align}
\min_{F,A_s,A_t} \mathcal{L}_{\mathrm{adv}} = 
    &-\mathbb{E}_{\mathbf{x}_{s} \sim P_{s}}[\log (1-D(A_s(F(\mathbf{x}_{s}))))] \nonumber \\
    &-\mathbb{E}_{\mathbf{x}_{t} \sim P_{t}}[\log D(A_t(F(\mathbf{x}_{t})))].
\label{eqn:adv_train}
\end{align}
Notably, only $\mathcal{L}_{\mathrm{adv}}$, which optimizes the feature extractor and the domain-specific attentions, is added to the overall objective function to ensure that the model is able to generate domain-invariant features.

\subsection{Dual Classifier based Iterative Self-Training}
With adversarial training, the distributions of source and target domains become globally aligned. However, the global alignment does not guarantee a good classification performance on the target domain, because of the difference in classification boundaries among source and target domains. Therefore, we propose a novel iterative self-training strategy to adjust the classification boundaries to fit the target domain and improve its classification performance.

Self-training converts the target domain predictions into pseudo labels and uses them to minimize the cross-entropy loss \cite{self_training_ref}. 
Given high-quality pseudo labels, they are treated as supervisory signals to adapt the decision boundaries of the classifier according to target domain classes.
However, due to the domain shift, finding high-quality target domain pseudo labels can be a challenging problem, and the generated ones might be noisy and inefficient, especially at the beginning of the training.
Nevertheless, we aim to \textit{first} minimize the number of incorrect pseudo labels and \textit{second} minimize the negative impact of these incorrect ones on the performance. To do so, we follow two main strategies.

First, we repeat the training of the model for $r$ iterations, where the pseudo labels generated in the previous iteration are used in the next one. Since the model will be very uncertain about the pseudo labels in the first iteration, we ignore the target classification loss at this iteration.
However, in the following iterations, we take it into consideration since the model becomes more confident about the pseudo labels after being trained to minimize the domain shift between source and target domains.
Second, we use dual classifiers $C_1$ and $C_2$ setup, which has two main benefits. First, it helps the model to avoid the variance in the training data. Second, the average prediction vector of the two classifiers decreases the probability of low-confident predictions.

Notably, we design the two classifiers such that they share the same architecture, i.e., a single fully connected layer. This helps limiting the total number of parameters in the model and avoid pruning to overfitting. Consequently, we need to ensure that their predictions are diversified and they do not converge to become the one classifier throughout training. Thus, we add a regularization term $ | \theta_{C_1}^\intercal \theta_{C_2} |$ on the weights of the two classifiers as inspired by \cite{tri_training}, where $\theta_{C_1}$, $\theta_{C_2}$ represent the weights of $C_1$ and $C_2$ respectively. This regularization term ensures the diversity of the two classifiers and helps them to produce different yet correct predictions.
The final prediction vector is the averaged vector of the predictions of both classifiers. 

Formally, in each iteration, we first calculate the average probability $\mathbf{p}_{t}$ of the two classifiers, and the corresponding target pseudo labels $\hat{y}_{t}$ as follows.
\begin{align}
    & \mathbf{p}_{t} = \frac{1}{2} \left[C_1(A_t(F(\mathbf{x}_{t})) + C_2(A_t(F(\mathbf{x}_{t}))\right], \label{eqn:pt} \\
    & \hat{y}_{t} = argmax(\mathbf{p}_{t}). \label{eqn:y_pseudo}
\end{align}

The target classification loss $\mathcal{L}_{\mathrm{cls}}^{t}$ based on the above pseudo labels is defined as follows.

\begin{align}
     \min_{F,A_t,C_1,C_2}  \mathcal{L}_{\mathrm{cls}}^{t}= -\mathbb{E}_{\mathbf{x}_{t} \sim P_{t}} \sum_{k=1}^K  \mathbbm{1}_{[\hat{y}_t = k]} \log \mathbf{p}_{t}^k,
\label{eqn:trg_cls}
\end{align}
where $\mathbbm{1}$ is the indicator function, which is set to be 1 when the condition is met, and set to 0 otherwise. The target classification loss $\mathcal{L}_{\mathrm{cls}}^{t}$ optimizes the feature extractor $F$, the target domain-specific attention $A_t$ as well as the dual classifiers $C_1$ and $C_2$.

Similarly, the source classification loss $\mathcal{L}_{\mathrm{cls}}^{s}$, which depends on the source labels $y_s$, is formalized as follows.
\begin{align}
    &\mathbf{p}_{s} = \frac{1}{2}~[C_1(A_s(F(\mathbf{x}_{s})) + C_2(A_s(F(\mathbf{x}_{s}))], \\
    &\min_{F,A_s,C_1,C_2} \mathcal{L}_{\mathrm{cls}}^{s}= -\mathbb{E}_{(\mathbf{x}_{s},y_{s}) \sim P_{s}} \sum_{k=1}^K  \mathbbm{1}_{[y_s = k]} \log \mathbf{p}_{s}^k , \label{eqn:src_cls}
\end{align}

where the source classification loss $\mathcal{L}_{\mathrm{cls}}^{s}$ optimizes the feature extractor $F$, the source domain-specific attention $A_s$ as well as the dual classifiers $C_1$ and $C_2$.

To sum up, we integrate the adversarial loss with the source and target classification losses and the regularization of the dual classifiers in one objective loss function as follows.

\begin{align}
    \mathcal{L}_{\mathrm{overall}} =  \mathcal{L}_{\mathrm{adv}} + \mathcal{L}_{\mathrm{cls}}^s + \lambda_1 \mathcal{L}_{\mathrm{cls}}^t + \lambda_2  | \theta_{C_1} ^\intercal   \theta_{C_2} |.
\label{eqn:overall}
\end{align}
Since the adversarial training and the source classification are two essential modules, we set their weights to one and tune the values of the two hyperparameters $\lambda_1$ and $\lambda_2$ to control their contributions.
In overall, the three losses are integrated to guide the feature extractor to generate domain-invariant features, while allowing the domain-specific attentions to preserve the key features for each domain. Additionally, the dual classifiers are diversified using the regularization term.
More details about the training can be found in Algorithm~\ref{algorithm}.

\begin{algorithm}
 \caption{Training procedure of our proposed ADAST framework.}
 \label{algorithm}
 \SetAlgoLined
\textbf{Input}: $X_s, X_t$, $D$, $F$, $\psi_s, \psi_t$, $\mathcal{C}_1$, $\mathcal{C}_2$\;
\textbf{Output}: Trained $F$, $\psi_t$ for the target domain\;
  1. Set $\lambda_1=0$\;
  2. \For{i=1 \textbf{to} r}{
      3. Sample mini-batches from source domain $(\mathbf{x}_s,y_s) \sim P_s$ and target domain $\mathbf{x}_t \sim P_t$\;
      4. Extract shared features using $F$, then unshared representations $\psi_s(F(\mathbf{x}_s))$, $\psi_t(F(\mathbf{x}_t))$\;
      5. Update $D$ by $\mathcal{L}_D$ (Eq.~\ref{eqn:train_disc})\;
      6. Update $F, \psi_s, \psi_t$ by $\mathcal{L}_{adv}$ (Eq.~\ref{eqn:adv_train})\;
      7. Update $F, \psi_s, \mathcal{C}_1, \mathcal{C}_2$ by $\mathcal{L}^s_{cls}$ (Eq.~\ref{eqn:src_cls})\;
      8. Penalize $\mathcal{C}_1$ and $\mathcal{C}_2$ weights similarity \;
      9. Generate pseudo labels $\hat{y}_{t}^{i}$ (Eq. \ref{eqn:pt},\ref{eqn:y_pseudo})\;
      10. Update $F, \psi_t, \mathcal{C}_1, \mathcal{C}_2$ by $\mathcal{L}^t_{cls}$ (Eq.~\ref{eqn:trg_cls})\;
      11. Set a value to $\lambda_1$\;
 }
\end{algorithm}

\section{Experiments}

\subsection{Datasets}
We evaluate the proposed framework on three challenging datasets, namely Sleep-EDF\footnote{https://physionet.org/physiobank/database/sleep-edfx/} (\textbf{EDF} for short), SHHS-1 (\textbf{S1}) and SHHS-2\footnote{https://sleepdata.org/datasets/shhs} (\textbf{S2}). These three datasets represent distinct domains due to their differences in sampling rates and EEG channels.

\subsubsection{Sleep-EDF (EDF)}
EDF dataset was first published in 2013 with the polysomnography (PSG) readings of 20 healthy subjects (10 males and 10 females). It contains data from two studies: Sleep Cassette (SC* files) and Sleep Telemetry (ST* files). The Sleep Cassette study (1987-1991) addresses the age effects on sleep, while the Sleep Telemetry study (1994) addressed the temazepam effects on sleep. Each PSG recording consists of two EEG channels namely Fpz-Cz and Pz-Oz, with a sampling rate of 100 Hz. We adopted the EEG recordings from Fpz-Cz channel following previous studies \cite{deepsleepnet,seqsleepnet,attnSleep_paper}.

\subsubsection{SHHS-1 (S1)}
SHHS \cite{shhs_ref1,shhs_ref2} is a multi-center cohort study conducted to  assess the cardiovascular and other consequences of sleep-disordered breathing. It tests the effect of some diseases such as stroke and hypertension on sleep-related breathing.
The data of S1 was recorded for a total of 6,441 men and women aged 40 years and older in their first visit, which was by the end of 1995 and lasted for two years.

\subsubsection{SHHS-2 (S2)}
The data of S2 represents the polysomnogram recordings of the second visit of 3,295 of the participants in S1. The outcome data was used to adjust the parent cohort.

Each PSG file in both S1 and S2 datasets contains data from two EEG channels namely C4-A1 and C3-A2, where we only adopt C4-A1 channel recordings for both datasets.
We selected subjects from S1 and S2 datasets such that 1) they contain different patients, 2) subjects from S2 dataset have a sampling rate of 250 Hz, and 3) the subjects have Apnea Hypopnea Index (AHI) $<1$ to eliminate the bias to sleep disorders and ensure a consistent clinical status of subjects ~\cite{AHI_reference}. Notably, we down-sampled the data from S1 and S2 datasets such that the sequence length is the same as the EDF dataset, i.e., 30 seconds $\times$ 100 Hz ($T=3,000$).

We preprocessed the three datasets by 1) merging stages N3 and N4 into one stage (N3) according to AASM standard, and 2) including only 30 minutes of wake stage periods before and after the sleep~\cite{deepsleepnet}. Table \ref{tbl:datasets} shows a brief summary of the above three datasets before down-sampling describing the number of subjects (\#Subjects) in each cross-domain, the selected EEG channel, the sampling rate, and the number of training (\#Train), validation (\#Val), and testing (\#Test) samples in each domain.

\subsection{Feature Extractor}
To extract the features from the EEG signals, we first preprocess the EEG signals to be split into 30-second segments (i.e., epochs). Each epoch is then passed through our feature extractor network to extract the features. 
We followed the design of the feature extractor proposed in \cite{emadeldeen_ts_tcc} which consists of three blocks, such that each block consists of a 1D-convolution layer, a batch normalization layer, a non-linear ReLU activation and a MaxPooling layer, as shown in Fig.~\ref{Fig:end-to-end}.
The 1D-convolution layer in the first block has 32 filters, with a kernel size of 25 and a stride of 6. The 1D-convolution layer in the second and third layers have 64 and 128 filters respectively and both have a kernel size of 8 and a stride of 1.  
The features extracted from these three blocks are then sent to the self-attention mechanism.

\begin{table}[!tb]
\caption{A brief description about the datasets.}
\centering
\resizebox{0.48\textwidth}{!}{
\begin{tabular}{@{}l|cccccc@{}}
\toprule
Dataset & \#Subjects & EEG Channel & Sampling Rate & \#Train & \#Val & \#Test \\ \midrule
EDF & 20 & Fpz-Cz & 100 & 25,612 & 7,786 & 8,910\\
S1 & 42 & C4-A1 & 125 & 24,515 & 7,948 & 9,067\\
S2 & 44 & C4-A1 & 250 & 31,613 & 9,769 & 12,413 \\ \bottomrule
\end{tabular}
}
\label{tbl:datasets}
\end{table}

\subsection{Experimental Settings}
To evaluate the performance of our model and baseline models, we employed the classification accuracy (ACC) and the macro-averaged F1-score (MF1). These two metrics are defined as follows:

\begin{align}
    &ACC = \frac{\sum_{i=1}^{K}TP_i}{M}, \label{equ:acc}\\
    &MF1 = \frac{1}{K} \sum_{i=1}^{K} \frac{2 \times Precision_i \times Recall_i}{Precision_i + Recall_i} \label{equ:f1},
\end{align}
where $Precision_i = \frac{TP_i}{TP_i + FP_i}$, and $ Recall_i = \frac{TP_i}{TP_i + FN_i} $. $TP_i, ~FP_i,~ TN_i$, and $FN_i$ denote the True Positives, False Positives, True Negatives, and False Negatives for the $i$-th class respectively, M is the total number of samples and K is the number of classes. All the experiments were repeated 5 times with different random seeds for model initialization, and then we reported the average performance (i.e., ACC and MF1) with standard deviation.

We performed \textit{subject-wise} splits for the data from the three domains, i.e., we split them into 60\%, 20\%, and 20\% for training, validation and testing, respectively, such that the data from one subject were assigned to either of the 3 splits. We used the training part of source and target domains while training our model. We used the validation part and test part of the target domain for validation and testing. Following \cite{tri_training,dirt}, we used the validation split of the target domain to select the best hyperparameters in our model. We tuned the parameters $\lambda_1, \lambda_2$ in the range $\{0.00001, 0.0001, 0.001, 0.01, 0.1, 1\}$, and set their values as $\lambda_1=0.01$ and $\lambda_2=0.001$. 
For iterative self-training, we set the maximum iterations $r$ to 2, as the performance of the model was found to converge.
We used Adam optimizer with a learning rate of 1e-3 that is decayed by 0.1 after 10 epochs, weight decay of 3e-4, $\beta_1 = 0.5$, $\beta_2 = 0.99$, and a batch size of 128. 
We trained the model for a predetermined number of epochs (15 epochs in our case) per iteration.
All the experiments were performed with PyTorch 1.7 on a NVIDIA GeForce RTX 2080 Ti GPU.
The source code and supplementary material are available at \href{https://github.com/emadeldeen24/ADAST}{https://github.com/emadeldeen24/ADAST}.

\newcommand{\STAB}[1]{\begin{tabular}{@{}c@{}}#1\end{tabular}}

\begin{table*}[!tb]
\centering
\caption{Comparison against various baselines. Best results are in bold, and the second best are underlined.}
\begin{tabular}{@{}l|l|cccccc|c@{}}
\toprule
& \multicolumn{1}{c|}{} & \multicolumn{6}{c|}{Cross-Domain Accuracy} & \multicolumn{1}{c}{AVG} \\ \midrule

& Baselines & EDF$\rightarrow$S1 & EDF$\rightarrow$S2 & S1$\rightarrow$EDF & S1$\rightarrow$S2 & S2$\rightarrow$EDF & S2$\rightarrow$S1 & ACC \\ \midrule

\multirow{4}{*}{\STAB{\rotatebox[origin=c]{90}{DT}}}

& DeepSleepNet \cite{deepsleepnet} & 49.19$\pm$3.09 & 48.48$\pm$2.74 & 76.37$\pm$0.11 & 61.53$\pm$0.82 & 62.45$\pm$0.26 & \textbf{78.97$\pm$0.11} & 62.83\\

& SleepEEGNet \cite{mousavi2019sleepeegnet} & 62.51$\pm$3.31 & 49.82$\pm$3.22 & 56.21$\pm$0.46 & 59.41$\pm$2.70 & 62.98$\pm$1.34 & 69.96$\pm$2.00  & 60.14 \\

& AttnSleep \cite{attnSleep_paper} & 64.44$\pm$3.37 & 57.43$\pm$5.45 & 75.41$\pm$0.65 & \underline{72.08$\pm$0.23} & 66.59$\pm$0.83 & 77.52$\pm$0.29 & 68.91\\

& Source-Only \textit{(Ours)} & 57.76$\pm$2.40 & 52.05$\pm$4.47 & 75.05$\pm$1.43 & 63.84$\pm$3.83 & 59.65$\pm$2.90 & 73.94$\pm$1.48 & 63.72  \\

\midrule


\multirow{8}{*}{\STAB{\rotatebox[origin=c]{90}{DA}}}

& Deep CORAL \cite{deep_coral}& 63.92$\pm$2.35 & 53.24$\pm$4.19 & 75.95$\pm$4.53 & 61.49$\pm$2.43 & 68.90$\pm$3.69 & 75.55$\pm$1.64 & 66.51  \\

& MDDA \cite{mdda} & 66.18$\pm$0.73 & 59.22$\pm$4.34 & 74.97$\pm$1.39 & 65.96$\pm$4.57 & 69.93$\pm$2.42 & 75.18$\pm$2.43 & 68.57 \\

& DSAN \cite{dsan}& 65.45$\pm$0.61 & \textbf{69.58$\pm$1.62} & \textbf{82.30$\pm$0.31} & 67.19$\pm$2.11 & 70.89$\pm$1.01 & 75.80$\pm$1.62 & \underline{71.87} \\  


& DANN \cite{dann} & 65.93$\pm$0.63 & 58.67$\pm$3.44 & 77.40$\pm$0.54 & 64.14$\pm$0.48 & 67.53$\pm$2.60 & 73.72$\pm$0.87 & 67.90 \\

& ADDA \cite{adda}& \underline{68.02$\pm$1.07} & 52.89$\pm$6.64 & \underline{80.73$\pm$1.50} & 57.85$\pm$0.72 &  72.65$\pm$0.28 & 71.63$\pm$1.10 & 67.30 \\

& CDAN \cite{cdan} & 64.15$\pm$2.41 & 64.09$\pm$1.27 & 78.02$\pm$0.84 & 66.06$\pm$2.29 & 72.13$\pm$2.13 & 76.42$\pm$1.88 & 70.14  \\

& DIRT-T \cite{dirt} & 66.51$\pm$2.99 & 59.20$\pm$4.34 & 79.98$\pm$0.35 & 65.06$\pm$3.98 & \underline{72.95$\pm$3.27} & 77.16$\pm$0.61 & 70.19  \\

& ADAST \textit{(Ours)} & \textbf{75.50$\pm$1.03} & \underline{67.56$\pm$2.37} & 75.94$\pm$1.25 & \textbf{72.27$\pm$1.06} & \textbf{75.28$\pm$1.78} & \underline{77.80$\pm$0.25} & \textbf{74.00} \\

\bottomrule
\toprule
& \multicolumn{1}{c|}{} & \multicolumn{6}{c|}{Cross-Domain F1-score} & \multicolumn{1}{c}{MF1} \\ \midrule

\multirow{4}{*}{\STAB{\rotatebox[origin=c]{90}{DT}}}

& DeepSleepNet \cite{deepsleepnet} & 42.43$\pm$2.85 & 39.93$\pm$2.37 & 64.58$\pm$0.44 & 55.02$\pm$0.61 & 53.22$\pm$0.22 & \textbf{66.65$\pm$0.65} & 53.63 \\

& SleepEEGNet \cite{mousavi2019sleepeegnet} & 55.29$\pm$3.09 & 40.68$\pm$4.63 & 51.97$\pm$0.77 & 55.10$\pm$1.10 & 54.74$\pm$1.42 & 61.04$\pm$2.98 & 53.13 \\

& AttnSleep \cite{attnSleep_paper} & 54.77$\pm$2.36 & 50.06$\pm$3.29 & 63.03$\pm$0.66 & \textbf{62.02$\pm$0.69} & 56.41$\pm$0.82 & \underline{63.71$\pm$0.12} & 58.33 \\

& Source-Only \textit{(Ours)} & 46.04$\pm$1.86 & 42.22$\pm$4.07 & 63.07$\pm$1.34 & 54.96$\pm$3.12 & 49.10$\pm$2.55 & 62.71$\pm$3.00 & 53.02  \\

\midrule

\multirow{8}{*}{\STAB{\rotatebox[origin=c]{90}{DA}}}

& Deep CORAL \cite{deep_coral} & 55.94$\pm$2.21 & 42.75$\pm$5.82 & 62.36$\pm$3.99 & 50.68$\pm$2.34 & 57.35$\pm$2.81 & 61.84$\pm$1.41 & 55.16 \\

& MDDA \cite{mdda} & 56.00$\pm$0.53 & 48.07$\pm$1.88 & 62.18$\pm$1.01 & 54.09$\pm$2.83 & 57.43$\pm$2.67 & 60.94$\pm$1.88 & 56.45  \\

& DSAN \cite{dsan}& 55.67$\pm$0.43 & \textbf{55.07$\pm$1.19} & \underline{67.78$\pm$0.28} & 55.55$\pm$1.16 & 58.79$\pm$1.04 & 62.20$\pm$0.93 & \underline{59.18}  \\ 


& DANN \cite{dann} & 54.79$\pm$0.54 & 48.49$\pm$2.57 & 63.86$\pm$0.69 & 53.48$\pm$0.36 & 57.14$\pm$2.08 & 60.17$\pm$0.65 & 56.32 \\

& ADDA \cite{adda} & \underline{58.18$\pm$1.41} & 43.96$\pm$6.26 & \textbf{68.43$\pm$0.88} & 48.53$\pm$0.64 & 59.02$\pm$0.54 & 58.54$\pm$0.78 & 56.11 \\

& CDAN \cite{cdan} & 52.61$\pm$2.31 & 52.42$\pm$0.33 & 64.31$\pm$0.93 & 54.73$\pm$1.45 & \underline{59.70$\pm$2.18} & 62.76$\pm$1.79 & 57.75 \\ 

& DIRT-T \cite{dirt} & 55.34$\pm$2.81 & 48.31$\pm$3.81 & 66.12$\pm$0.45 & 54.55$\pm$2.85 & 57.72$\pm$4.61 & 62.05$\pm$0.72 & 57.35 \\

& ADAST \textit{(Ours)} & \textbf{61.92$\pm$0.83} & \underline{53.80$\pm$2.11} & 63.33$\pm$1.02 & \underline{58.69$\pm$0.60} & \textbf{62.49$\pm$1.04} & 63.10$\pm$0.06 & \textbf{60.39} \\

\bottomrule
\end{tabular}
\label{tbl:baselines_comparison}
\end{table*}

\subsection{Baselines}
To assess our proposed ADAST model, we compared it against various baselines. We first included the Direct Transfer (DT) results of three sleep staging methods. These methods are \textbf{DeepSleepNet} \cite{deepsleepnet}; \textbf{SleepEEGNet} \cite{mousavi2019sleepeegnet}; and \textbf{AttnSleep} \cite{attnSleep_paper} (refer to Section \ref{subsec:ssc} for their details). In addition, we adopted seven state-of-the-art discrepancy- and adversarial-based domain adaptation (DA) baselines. In particular, Deep CORAL, MDDA and DSAN are discrepancy-based methods, while DANN, ADDA, CDAN, and DIRT-T are adversarial-based methods. These baselines are summarized as follows.

\begin{itemize}
    \item \textbf{Deep CORAL} \cite{deep_coral}: it extends CORAL \cite{coral} to learn a nonlinear transformation that aligns the correlations of layer activations in deep neural networks.
    
    \item \textbf{MDDA} \cite{mdda}: it applies MMD and CORAL on multiple classification layers to minimize the discrepancy between the source and target domains.
    
    \item \textbf{DSAN} \cite{dsan}: it incorporates a local MMD loss to align the same-class sub-domain distributions. 
    
    \item \textbf{DANN} \cite{dann}: it jointly trains feature extractor and domain classifier by negating the gradient from the domain classifier with a gradient reversal layer (GRL). 
    
    \item \textbf{ADDA} \cite{adda}: it performs a similar operation as DANN but by inverting the labels instead of using GRL.
    
    \item \textbf{CDAN} \cite{cdan}: it minimizes the cross-covariance between feature representations and classifier predictions.
    
    \item \textbf{DIRT-T} \cite{dirt}: it combines virtual adversarial domain adaptation with a teacher model to refine the decision boundary of the target domain. 
    
\end{itemize}

Notably, we included the results of the \textbf{Source-Only} experiment, which refers to the DT results of our backbone network.
In addition, we used our backbone feature extractor for all the seven DA baselines to ensure a fair comparison. We tuned the hyperparameters of the baselines to achieve their best performance.

\subsection{Experimental Results}
\label{sec:exp_results}

Table~\ref{tbl:baselines_comparison} shows the comparison results among various methods. Overall, the direct transfer usually achieves lower performance than domain adaptation. The results of DT experiments on \cite{deepsleepnet,mousavi2019sleepeegnet,attnSleep_paper} indicate that the domain shift problem causes a big performance drop and should be addressed separately. Therefore, it becomes important to use domain adaptation to address the domain shift problem for cross-dataset sleep stage classification, which is supported by the results of the other seven DA baselines.

It should be highlighted that the performance increase in the seven DA baselines should be compared to our Source-Only DT results, as they share the same backbone network.
We noticed that the three methods considering the class-conditional distribution, i.e., CDAN, DIRT-T, and DSAN, outperform the ones globally aligning the source and target domains, i.e., DANN, Deep CORAL, ADDA, and MDDA. This indicates that considering class distribution, especially in the case of imbalanced sleep data, is important to achieve better classification performance on the target domain. 
Our proposed ADAST achieves superior performance over all the baselines in terms of both mean accuracy and F1-score in four out of six cross-domain scenarios for two reasons. First, our ADAST, similar to CDAN, DIRT-T, and DSAN, also considers the class-conditional distribution. In particular, ADAST explores the target domain classes using the proposed iterative self-training strategy with dual classifiers. Second, ADAST preserves domain-specific features using the unshared attention module, which improves the performance. 

As shown in Table \ref{tbl:baselines_comparison}, the performance of our model is less than most baselines in the scenario S1$\rightarrow$EDF. Note that we used the same value of $\lambda_1$ (i.e., 0.01) for all the six scenarios, which might not be fair for some scenarios. We found that the quality of the pseudo labels is not good in this scenario S1$\rightarrow$EDF, and thus we should use a smaller $\lambda_1$ to reduce the contribution of the target classification loss. By tuning $\lambda_1$ from 0.01 to $10^{-6}$, the mean accuracy and MF1 of our ADAST in the scenario S1$\rightarrow$EDF would increase from 75.94\% and 63.33\% to 78.50\% and 64.73\%, respectively. Please refer to Fig. S.1c in the supplementary material for more details.

\begin{table*}
\centering
\caption{Ablation study showing the different variants of our proposed ADAST framework. \textbf{ATT}: domain-specific ATTention, \textbf{DC}: Dual Classifiers, \textbf{ST}: Self Training. The first row indicates using only the domain discriminator along with a single classifier.}
\begin{tabular}{@{}ccc|cccccc|cc@{}}
\toprule
\multicolumn{3}{c|}{Component} & \multicolumn{6}{c|}{Cross-Domain Accuracy} & \multicolumn{1}{c}{AVG} \\ \midrule

ATT & DC & ST & EDF$\rightarrow$S1 & EDF$\rightarrow$S2 & S1$\rightarrow$EDF & S1$\rightarrow$S2 & S2$\rightarrow$EDF & S2$\rightarrow$S1 & ACC \\ \midrule

- & - & - & 66.46$\pm$1.86 & 62.72$\pm$4.14 & \textbf{79.10$\pm$1.23} & 61.54$\pm$0.08 & 72.28$\pm$3.33 & 73.19$\pm$0.64 & 69.21  \\

\checkmark  & - & -  & 71.31$\pm$1.78 & \textbf{69.07$\pm$1.50} & 76.83$\pm$1.83 & 63.98$\pm$3.11 & 74.59$\pm$0.93 & 75.19$\pm$1.63 & 71.50 \\
 
\checkmark  & \checkmark & -  & 72.53$\pm$0.88 & 66.50$\pm$2.45 & 78.22$\pm$0.48 & 68.54$\pm$5.29 & \textbf{75.57$\pm$0.86} & 76.98$\pm$0.54 & 73.05  \\

\checkmark & - & \checkmark & 70.76$\pm$2.22 & 68.79$\pm$0.20 & 77.61$\pm$0.79 & 68.58$\pm$5.14 & 74.05$\pm$1.73 & 75.82$\pm$0.67 & 72.60 \\

\checkmark & \checkmark   & \checkmark & \textbf{75.50$\pm$1.03} & 67.56$\pm$2.37 & 75.54$\pm$1.25 & \textbf{72.27$\pm$1.06} & 75.28$\pm$1.78 & \textbf{77.80$\pm$0.25} & \textbf{74.00} \\ 
\bottomrule

\end{tabular}
\label{tbl:ablation}
\end{table*}

We also observed interesting results while investigating different cross-dataset scenarios. Various methods usually achieve better performance in the cross-domain scenario S1$\rightarrow$S2 than EDF$\rightarrow$S2 (and similarly S2$\rightarrow$S1 is better than EDF$\rightarrow$S1).
To explain this, as shown in Table \ref{tbl:datasets}, S1 and S2 are closer to each other, as they have the same EEG channel.
Meanwhile, EDF has a different EEG channel and sampling rate, and thus it is a distant domain from S1 and S2.
These results indicate that distant domain adaptation is still very challenging. 
Finally, we observed that S1$\rightarrow$EDF is easier than S2$\rightarrow$EDF, probably because S1 and EDF have close sampling rates to each other. 

\begin{figure}[!tb] 
  \begin{subfigure}[b]{0.5\linewidth}
    \centering
    \includegraphics[width=\linewidth]{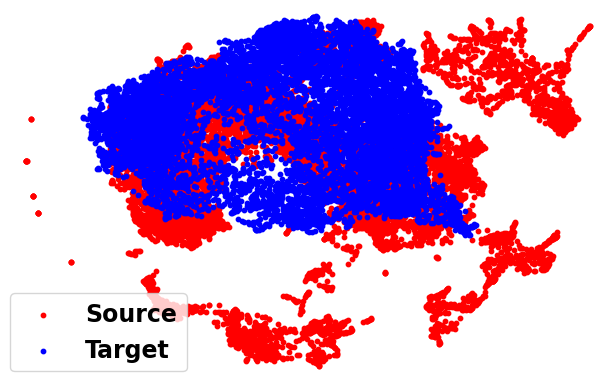}  
    \caption{} 
    \label{fig:src_trg_alignment:a} 
  \end{subfigure}
  \begin{subfigure}[b]{0.5\linewidth}
    \centering
    \includegraphics[width=\linewidth]{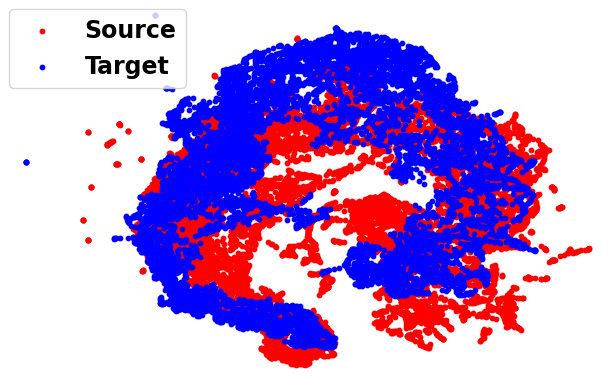} 
    \caption{} 
    \label{fig:src_trg_alignment:b} 
  \end{subfigure} 
  \caption{UMAP feature space visualization showing the source and target domains alignment using (a) Source-Only, and (b) our ADAST, applied for the scenario S2$\rightarrow$EDF.}
  \label{fig:src_trg_alignment} 
\end{figure}

\begin{figure}[!tb] 
  \begin{subfigure}[b]{0.5\linewidth}
    \centering
    \includegraphics[width=\linewidth]{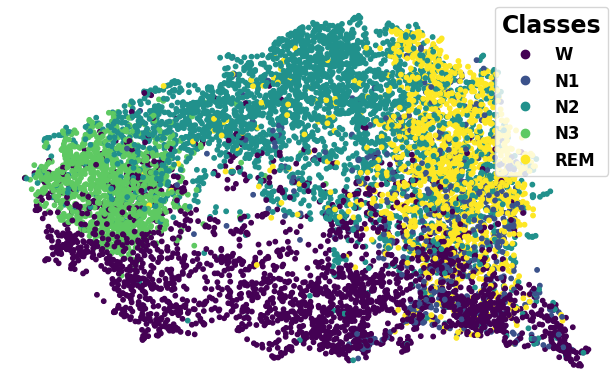}  
    \caption{} 
    \label{fig:trg_classification:a} 
  \end{subfigure}
  \begin{subfigure}[b]{0.5\linewidth}
    \centering
    \includegraphics[width=\linewidth]{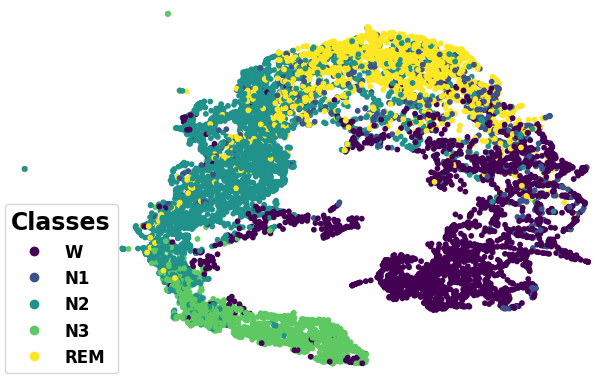}  
    \caption{} 
    \label{fig:trg_classification:b} 
  \end{subfigure} 
  \caption{UMAP feature space visualization showing the target domains classification performance after (a) Source-Only, and (b) our ADAST alignment, applied for the scenario S2$\rightarrow$EDF.}
  \label{fig:trg_classification} 
\end{figure}

\subsection{Ablation Study}
We assessed the contribution of each component in our ADAST framework, namely the unshared domain-specific attention module (\textbf{ATT}), the dual classifiers (\textbf{DC}) and self-training (\textbf{ST}). Particularly, we conducted an ablation study to show the results of different variants of ADAST in Table~\ref{tbl:ablation}. 

The results emphasize three main conclusions. First, using the proposed domain-specific attention benefits the overall performance, as it helps to preserve the domain-specific features.
Second, the self-training improves the classification performance by $\sim$ 1.1\%. This improvement shows the benefit of incorporating the target domain class information in modifying the classification boundaries by using pseudo labels.
Third, the addition of dual classifiers benefits the classification performance in overall as it avoids the variance in the training data. Moreover, combining it with the self-training in specific is helpful to further improve the performance by 2.5\% through improving the quality of the pseudo labels.

\subsection{Representation Visualization}
In Section \ref{sec:exp_results}, the results illustrate the advantages of our proposed ADAST framework over the initial Source-Only performance.
To make the comparison more intuitive, we visualized the feature representations that are learned during the training process using Uniform Manifold Approximation and Projection (UMAP) \cite{umap}. 

First, we investigated the alignment quality, where Fig.~\ref{fig:src_trg_alignment} visualizes the source and target alignment in the scenario S2$\rightarrow$EDF. In particular, Fig.~\ref{fig:src_trg_alignment:a} shows the Source-Only alignment, and Fig.~\ref{fig:src_trg_alignment:b} shows our ADAST framework alignment.
In these figures, the red dots represent the source domain, and the blue dots denote the target domain. We can observe that the Source-Only is not very efficient as many disjoint patches are not well-aligned with the target domain. However, our ADAST framework improves the alignment of the two domains to become arc-shaped, which increases the overlapped region and they become less discriminative.

Additionally, we investigated the target domain classification performance in the aforementioned scenario after the alignment in Fig.~\ref{fig:trg_classification}. In particular, Fig.~\ref{fig:trg_classification:a} is the Source-Only performance, and Fig.~\ref{fig:trg_classification:b} is the one after our alignment.
We noticed that the Source-Only alignment generates a lot of overlapping samples from different classes, which degrades the target domain classification performance. On the other hand, our ADAST framework improves the discrimination between the classes and they become more distinct from each other. This is achieved with the aid of the iterative self-training strategy.

\subsection{Domain-Specific vs Domain-Invariant Features}
In this subsection, we show that domain-invariant features has been extracted in two different manners.

\emph{Quantitatively:}
To have a quantitative insight about extracting domain-specific information, we use the Kullback-Leibler (KL) divergence as a distribution similarity measure.
In general, KL-divergence is a non-symmetric measure of the difference between two probability distributions $p(x)$ and $q(x)$.
Specifically, the KL-divergence of $q(x)$ from $p(x)$, denoted as  $D_{KL}(p(x), q(x))$, is a measure of the information lost when $q(x)$ is used to approximate $p(x)$ \cite{kl_ref}.
So, if we can approximate $p(x)$ with $q(x)$ without a big loss in information (achieve a low KL divergence value), then we can conclude more similarity between the two distributions, and vice versa.

To validate this idea, we first extract the reference features for each domain. To do so, we use the feature extractor network and the dual-classifiers to train the source domain data with the cross-entropy loss, and freeze the extracted features from the last epoch (i.e., $R_{s}$). Since we do not have access to the target domain labels, we use the pretrained model (on source domain data) to obtain the target domain features (i.e., $R_{t}$).

Next, we train the original model with source and target data to obtain the domain-invariant (DI) and domain-specific (DS) features before and after the domain-specific attention module respectively.
Finally, we calculate the following KL-divergence on the source domain sides as:  $D_{KL}(DI_{s}, R_{s})$ and $D_{KL}(DS_{s}, R_{s})$.
Similarly, we calculate the following KL-divergence on the target domain sides as:  $D_{KL}(DI_{t}, R_{t})$ and $D_{KL}(DS_{t}, R_{t})$.

We perform this experiment on the cross-domain scenario (EDF$\rightarrow$S1), and provide the results in Table~\ref{tbl:kl_div}.
Notably, the KL-divergence value is the minimum between the domain-specific features and their corresponding original domain features. This indicates a minimal information loss when trying to approximate the domain-specific features using the reference features, i.e., more similarity between them.
Therefore, we conclude the efficacy of the domain-specific attention module in preserving the unique characteristics of each domain.

\begin{table}[!htb]
\centering
\caption{KL-divergence between original features (R) and domain-invariant (DI) and domain-specific (DS) features applied for both source and target domains.}
\begin{tabular}{@{}c|c|cc@{}}
\toprule
&     & Domain-Invariant (DI)     & Domain-Specific (DS)     \\ \midrule
Source domain & $R_s$ & 0.7591 & 0.7346 \\ \midrule
Target domain & $R_t$ & 0.7849 & 0.6735 \\ \bottomrule
\end{tabular}
\label{tbl:kl_div}
\end{table}

\begin{figure*}[!tb] 
  \begin{subfigure}[b]{0.24\linewidth}
    \centering
    \includegraphics[width=\linewidth]{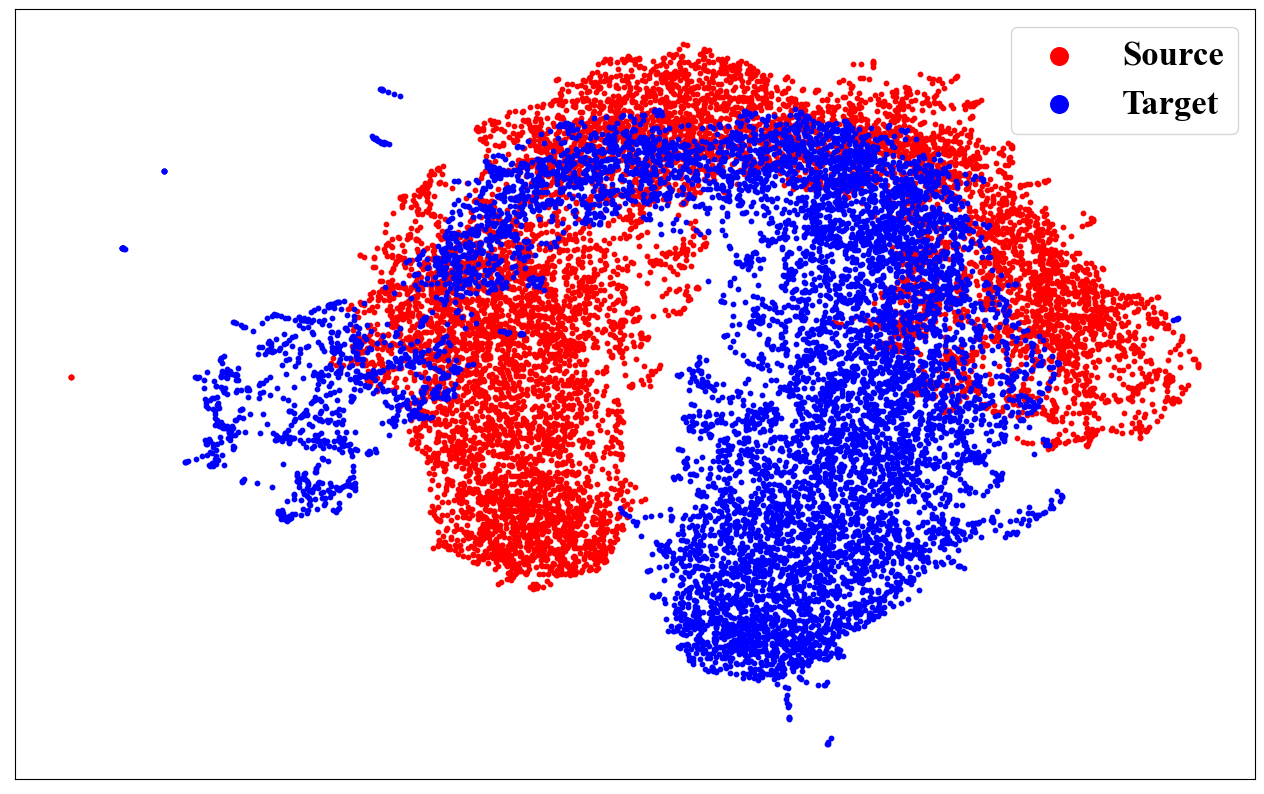}  
    \caption{} 
    \label{fig:domains_DI} 
  \end{subfigure}
  \begin{subfigure}[b]{0.24\linewidth}
    \centering
    \includegraphics[width=\linewidth]{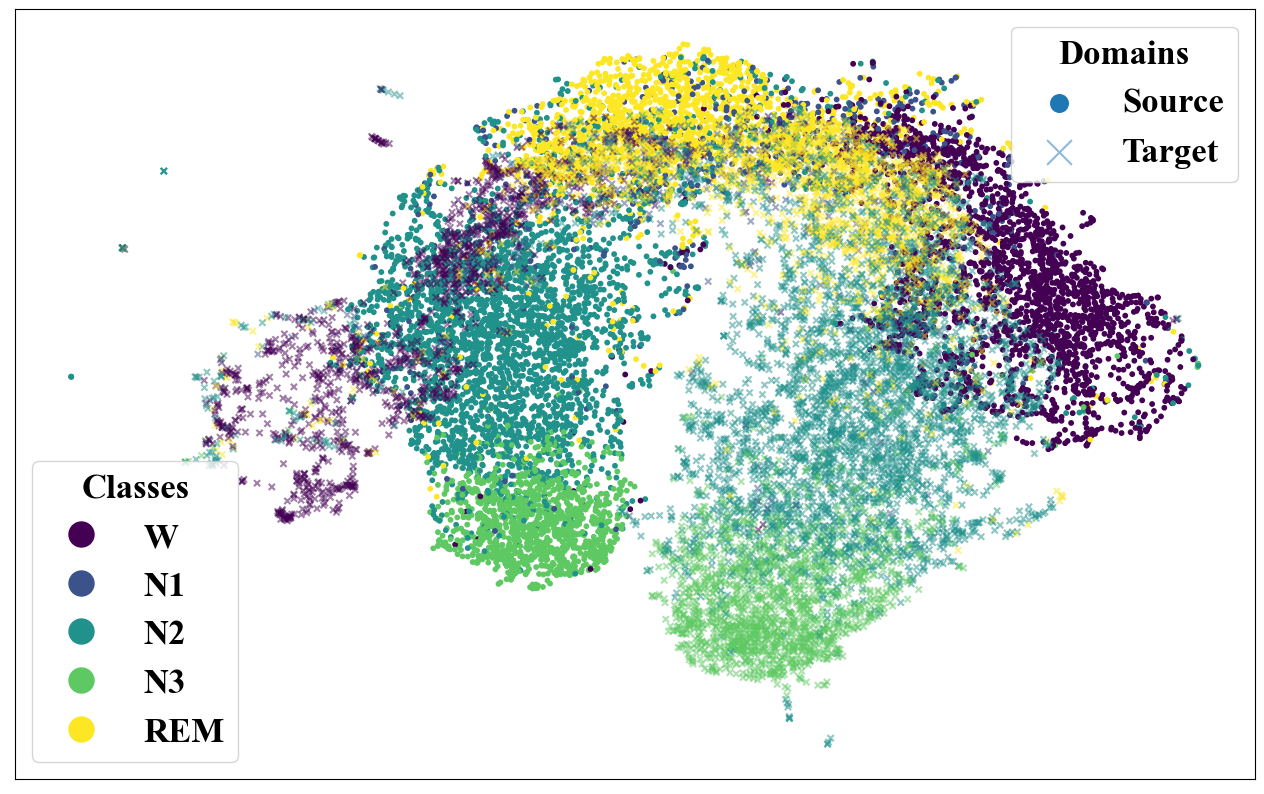}  
    \caption{} 
    \label{fig:classes_DI} 
  \end{subfigure} 
\begin{subfigure}[b]{0.24\linewidth}
    \centering
    \includegraphics[width=\linewidth]{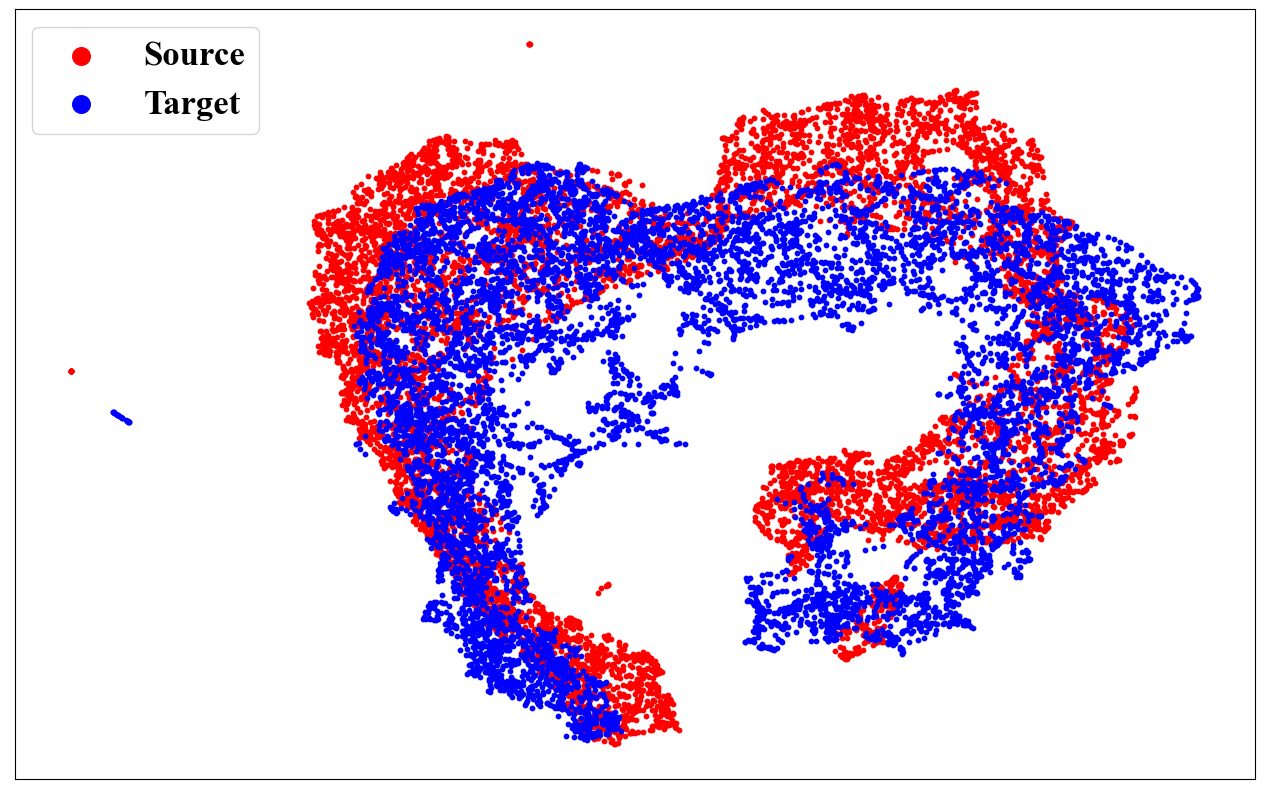}  
    \caption{} 
    \label{fig:domains_DS} 
  \end{subfigure} 
\begin{subfigure}[b]{0.24\linewidth}
    \centering
    \includegraphics[width=\linewidth]{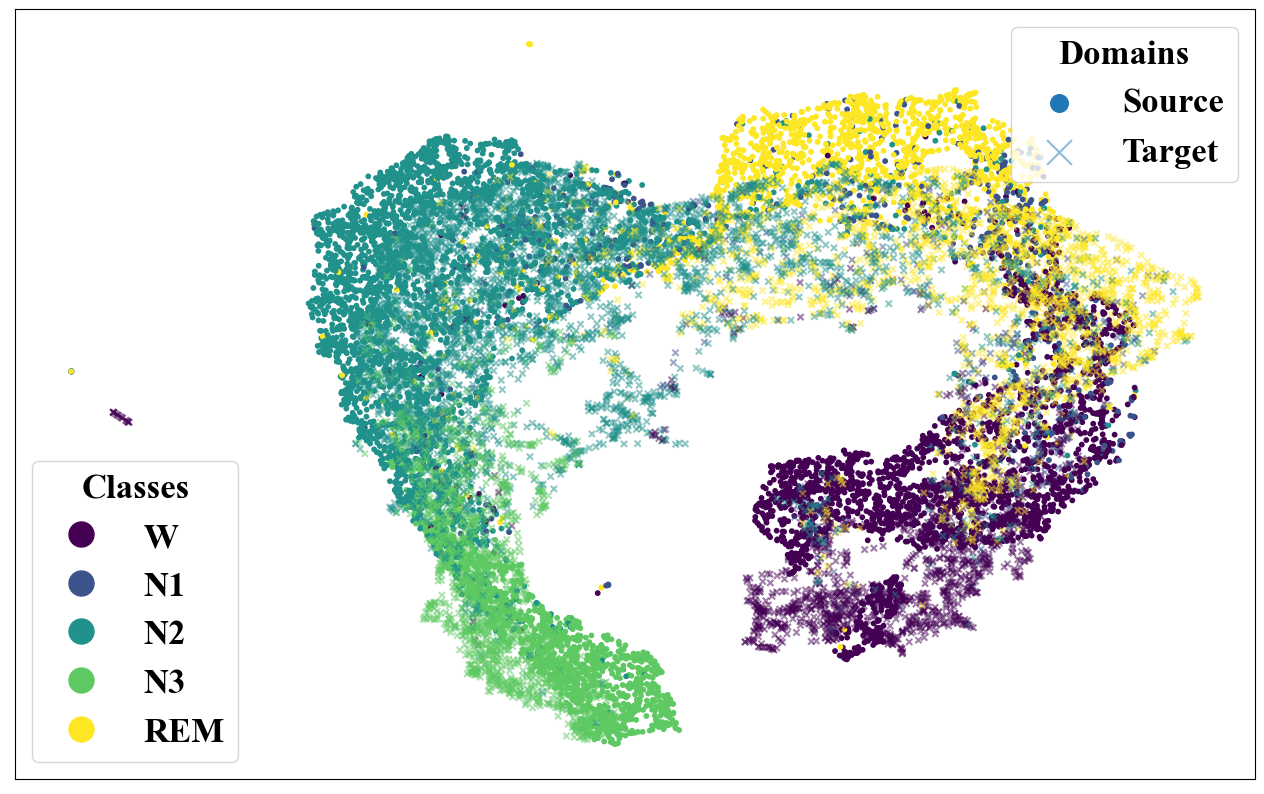}  
    \caption{} 
    \label{fig:classes_DS} 
  \end{subfigure} 
  \caption{UMAP feature space visualization showing the domain-invariant alignment (a) domain-wise , and (b) class-wise, besides the domain-specific alignment (c) domain-wise, and (d) class-wise  under the scenario of EDF$\rightarrow$S1.}
  \label{fig:visuals} 
\end{figure*}

\emph{Visual Inspection}
To analyze the feature space of both domain-invariant and domain-specific features, we visualize their distribution in the scenario (EDF$\rightarrow$S1) with UMAP, as shown in Fig.~\ref{fig:visuals}. 
We notice that the feature extractor can extract domain-invariant information by minimizing the distance between source and target distributions as shown in Fig.~\ref{fig:domains_DI}. 
However, these domain-invariant features still mis-classify some classes as seen in Fig.~\ref{fig:classes_DI}.
Meanwhile, the domain-specific attention helps to better align the domains (Fig.~\ref{fig:domains_DS}), as well as improving the class-wise alignment as in Fig.~\ref{fig:classes_DS}.
In addition, the wake class (Magenta) is now closer to the Rapid Eye Movement (REM) class (Yellow), which is reasonable since both classes share related information. Similarly, the points of N3 class become closer to N2 class. This implies that our model well learns specific features that can be adapted to the new unseen domain.

\subsection{Sensitivity Analysis}
\begin{figure}[!tb]
    \centering
    \includegraphics[width=0.7\columnwidth]{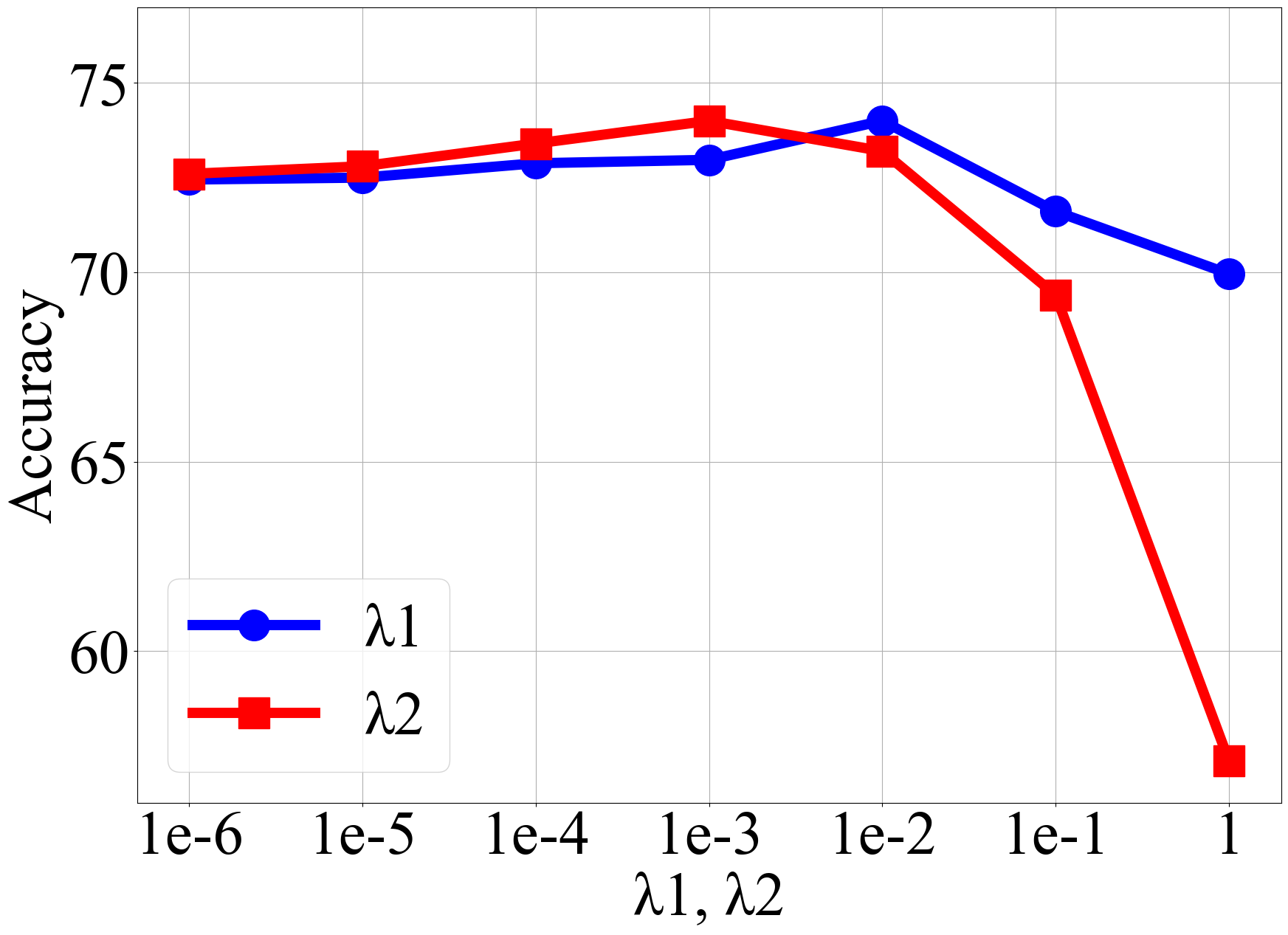}
    \caption{Sensitivity analysis to the different variants of $\lambda_1$ and $\lambda_2$ in Eq.\ref{eqn:overall}.}
    \label{fig:sens_analysis}
\end{figure}

\textbf{Effect of target classification loss.} 
Since the self-training process relies on target domain pseudo labels, it is not practical to assign a high weight to the target classification loss as the pseudo labels are expected to have some uncertainties.
Therefore, we studied the effect of the different variants to the weight assigned to the target classification loss $\lambda_1$, as shown in Fig.~\ref{fig:sens_analysis}. 

Notably, when $\lambda_1$ is very small (i.e., $\lambda_1$ = 1e-6), it makes the self-training useless, and the performance becomes very close to the case without self-training. As we gradually increase $\lambda_1$ value, we notice improvement in the overall performance until we reach the optimal value of $\lambda_1=0.01$. Further increasing $\lambda_1$ deteriorates the performance as the model is highly penalized based on the pseudo labels which may contain false examples.

\textbf{Effect of classifier weight constraint.} 
Since the dual classifiers share the same architecture, it is important to keep their predictions relatively different but not with a big gap.
The classifier weight constraint is the factor that keeps this distance with an acceptable margin, and hence, it becomes important to study the effect of this term and how its weight $\lambda_2$ should be selected.
We analyzed the performance of our model with different $\lambda_2$ values, as illustrated in Fig.~\ref{fig:sens_analysis}.

When $\lambda_2$ is very small, it makes the two classifiers perform very closely to each other, which has a similar performance with a single classifier. The performance is gradually improved when increasing $\lambda_2$, as the two classifiers tend to have different classification decisions. It can be found that the best performance is achieved with $\lambda_2=0.001$.
However, as its value is increased beyond this threshold (i.e., 0.001), we notice that the overall performance degrades. This happens as the weights of the two classifiers became very dissimilar, moving them away from the correct predictions.

\section{Conclusions}

In this paper, we proposed a novel adversarial domain adaptation architecture for sleep stage classification using single-channel raw EEG signals. We tackle the problem of the domain shift that happens when training the model on one dataset (i.e., the source domain), and testing it on another out of distribution dataset (i.e., the target domain). We developed unshared attention mechanisms to preserve domain-specific features. We also proposed a dual classifier-based iterative self-training strategy, which helps the model to adapt the classification boundaries according to the target domain with robust pseudo labels. The experiments performed on six cross-domain scenarios generated from three public datasets prove that our model can achieve superior performance over state-of-the-art domain adaptation methods. Additionally, the ablation study shows that the dual classifier-based self-training is the main contributor to the improvement as it considers class-conditional distribution in the target domain.

\bibliographystyle{unsrt}
\bibliography{citations}

\end{document}


\title{Supplementary material of ``ADAST: Attentive Cross-domain EEG-based Sleep Staging Framework with Iterative Self-Training"}
\maketitle

\renewcommand\thefigure{S.\arabic{figure}}
\renewcommand{\thesection}{S.\Roman{section}} 
\renewcommand\thetable{S.\arabic{table}}
\setcounter{figure}{0}
\setcounter{table}{0}
\setcounter{section}{0}

\section{Sensitivity Analysis}
\begin{figure*}[b!]
    \centering 
\begin{subfigure}{0.25\textwidth}
  \includegraphics[width=\linewidth]{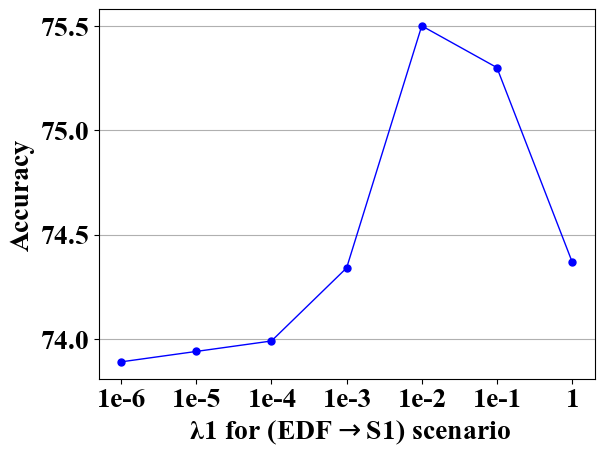}
  \caption{}
  \label{fig:sens_trg_cls_ab}
\end{subfigure}\hfil 
\begin{subfigure}{0.25\textwidth}
  \includegraphics[width=\linewidth]{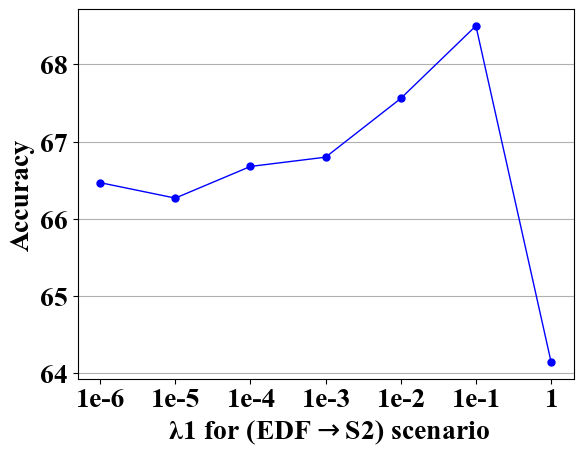}
  \caption{}
  \label{fig:sens_trg_cls_ac}
\end{subfigure}\hfil 
\begin{subfigure}{0.25\textwidth}
  \includegraphics[width=\linewidth]{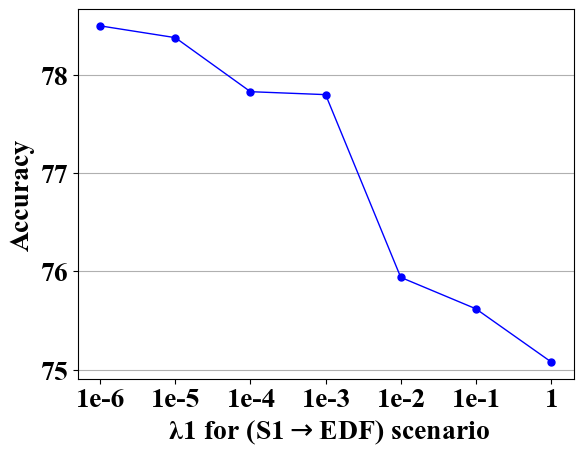}
  \caption{}
  \label{fig:sens_trg_cls_ba}
\end{subfigure}

\medskip
\begin{subfigure}{0.25\textwidth}
  \includegraphics[width=\linewidth]{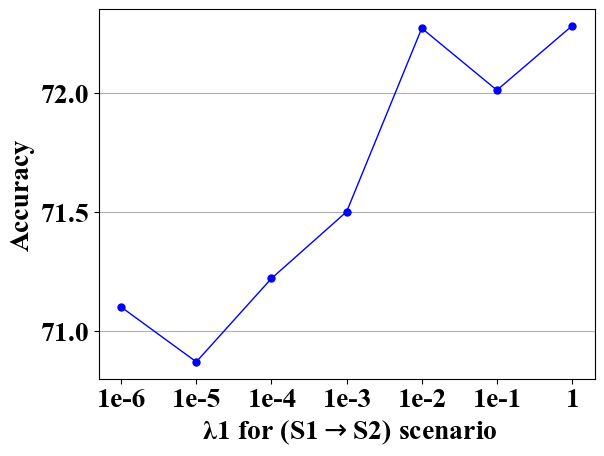}
  \caption{}
  \label{fig:sens_trg_cls_bc}
\end{subfigure}\hfil
\begin{subfigure}{0.25\textwidth}
  \includegraphics[width=\linewidth]{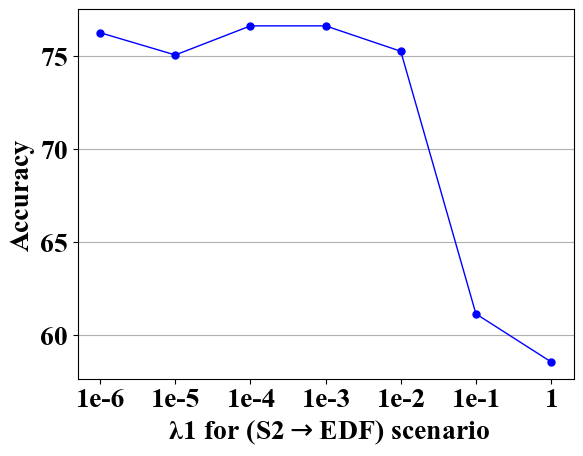}
  \caption{}
  \label{fig:sens_trg_cls_ca}
\end{subfigure}\hfil
\begin{subfigure}{0.25\textwidth}
  \includegraphics[width=\linewidth]{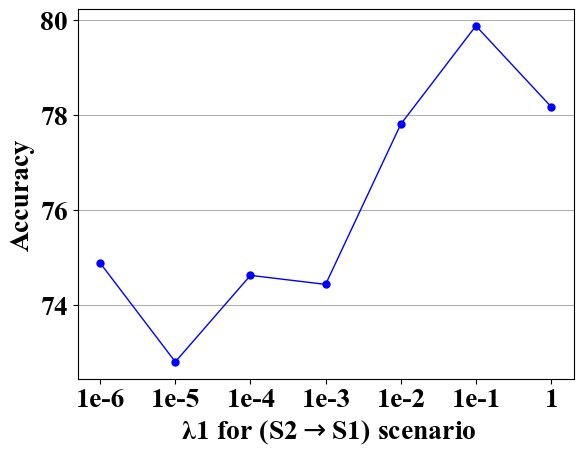}
  \caption{}
  \label{fig:sens_trg_cls_cb}
\end{subfigure}
\caption{Sensitivity analysis to the effect of different variants of $\lambda_1$ on the six cross-domain scenarios.}
\label{fig:lambda1_cross_domains}
\end{figure*}

We study the different variants of $\lambda_1$ in each of the cross domain scenarios. Fig. \ref{fig:sens_trg_cls_ba}.

